\documentclass[sigconf]{acmart}

\AtBeginDocument{%
  }

\usepackage{graphicx}
\usepackage{multirow}
\usepackage{svg}
\usepackage{booktabs}
\usepackage{array}
\usepackage{float}
\usepackage{hyperref}

\copyrightyear{2024}
\acmYear{2024}
\setcopyright{acmlicensed}\acmConference[ICMI Companion
'24]{INTERNATIONAL CONFERENCE ON MULTIMODAL INTERACTION}{November
4--8, 2024}{San Jose, Costa Rica}
\acmBooktitle{INTERNATIONAL CONFERENCE ON MULTIMODAL INTERACTION
(ICMI Companion '24), November 4--8, 2024, San Jose, Costa Rica}
\acmDOI{10.1145/3686215.3688381}
\acmISBN{979-8-4007-0463-5/24/11}

\begin{document}

\title{Gaze-Informed Vision Transformers: Predicting Driving Decisions Under Uncertainty}

\author{Sharath Koorathota}
\authornote{Both authors contributed equally to this research}
\authornote{Corresponding author}
\affiliation{%
  \institution{Columbia University}
  \city{New York}
  \country{USA}}
\email{sk4172@columbia.edu}

\author{Nikolas Papadopoulos}
\authornotemark[1]
\affiliation{%
  \institution{Columbia University}
  \city{New York}
  \country{USA}}
\email{np2832@columbia.edu}

\author{Jia Li Ma}
\affiliation{%
  \institution{Columbia University}
  \city{New York}
  \country{USA}}
\email{jm5299@columbia.edu}

\author{Shruti Kumar}
\affiliation{%
  \institution{Columbia University}
  \city{New York}
  \country{USA}}
\email{sk5111@columbia.edu}

\author{Xiaoxiao Sun}
\affiliation{%
  \institution{Columbia University}
  \city{New York}
  \country{USA}}
\email{xs2362@columbia.edu}

\author{Arunesh Mittal}
\affiliation{%
  \institution{Columbia University}
  \city{New York}
  \country{USA}}
\email{arunesh.mitl@gmail.com}

\author{Patrick Adelman}
\affiliation{%
  \institution{Georgia Institute of Technology}
  \city{Atlanta}
  \country{USA}}
\email{adelmanp@gmail.com}

\author{Paul Sajda}
\affiliation{%
  \institution{Columbia University}
  \city{New York}
  \country{USA}}
\email{psajda@columbia.edu}

\renewcommand{\shortauthors}{Koorathota, Papadopoulos et al.}

\begin{abstract}
Vision Transformers (ViT) have advanced computer vision, yet their efficacy in complex tasks like driving remains less explored. This study enhances ViT by integrating human eye gaze, captured via eye-tracking, to increase prediction accuracy in driving scenarios under uncertainty in both real-world and virtual reality scenarios. First, we establish the significance of human eye gaze in left-right driving decisions, as observed in both human subjects and a ViT model. By comparing the similarity between human fixation maps and ViT attention weights, we reveal the dynamics of overlap across individual heads and layers. This overlap demonstrates that fixation data can guide the model in distributing its attention weights more effectively. We introduce the fixation-attention intersection (FAX) loss, a novel loss function that significantly improves ViT performance under high uncertainty conditions. Our results show that ViT, when trained with FAX loss, aligns its attention with human gaze patterns. This gaze-informed approach has significant potential for driver behavior analysis, as well as broader applications in human-centered AI systems, extending ViT’s use to complex visual environments.
\end{abstract}



\begin{teaserfigure}
    \centering
    \includegraphics[width=0.8\textwidth]{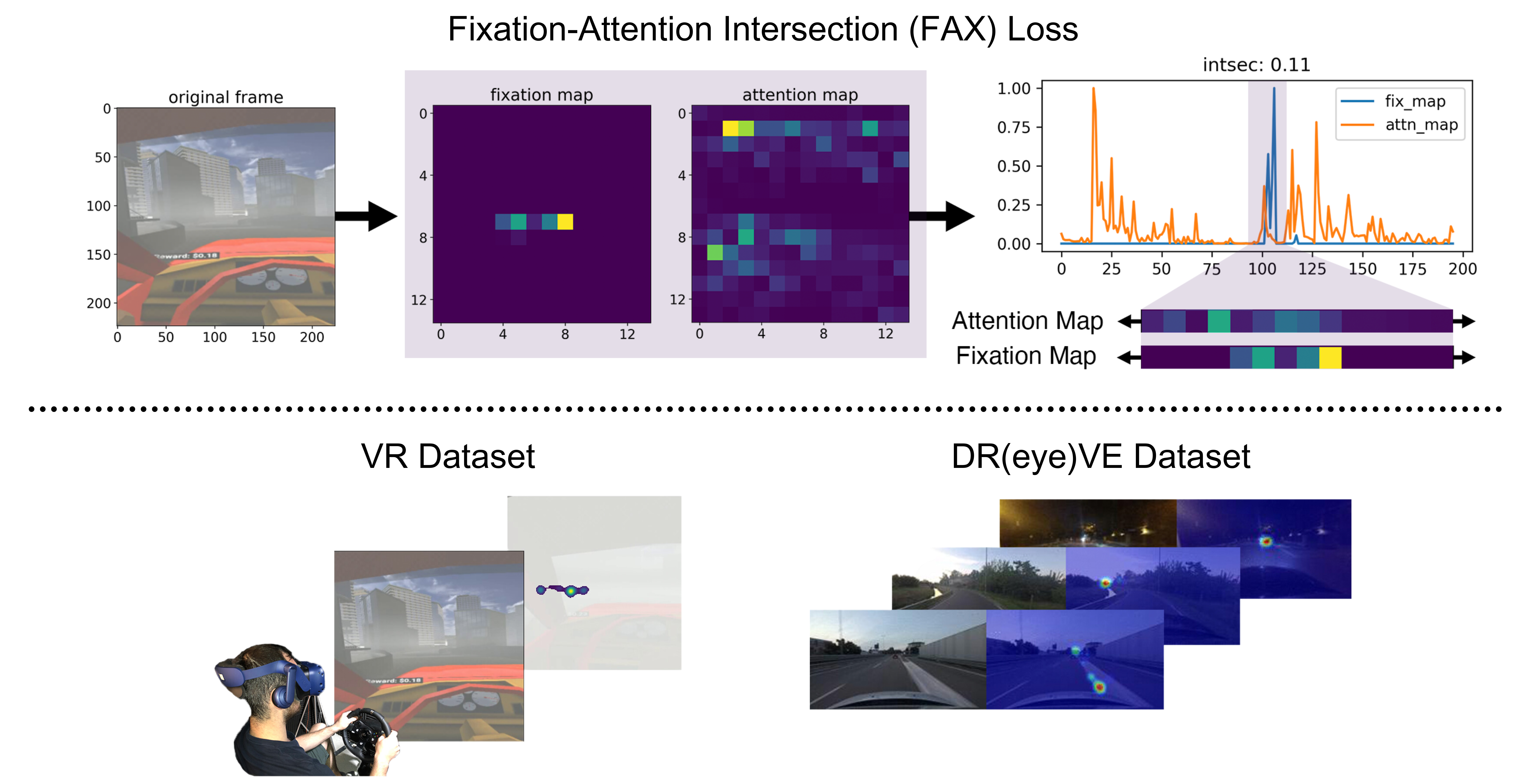}
    \caption{Our method integrates human gaze into Vision Transformers (ViT) to improve the prediction of driving decisions (left-right turns). We propose training ViT with a modified loss function, Fixation-Attention Intersection (FAX) loss, which calculates the intersection (dot product) of the model's attention map with the human fixation map. Experiments with both virtual reality and real-world datasets demonstrate that gaze integration enhances ViT accuracy in uncertain conditions (e.g., fog, bad weather/light conditions). When trained with FAX loss, the model's attention aligns with human gaze. Additionally, we release a novel dataset of human driving decisions collected in virtual reality to study turn behavior.}
    \label{fig:herofig}
\end{teaserfigure}


\maketitle

\section{Introduction}
\label{sec:intro}

The performance of Vision Transformers (ViT) \cite{dosovitskiy2020}, has exceeded human performance across various visual tasks. ViT have exhibited state-of-the-art performance in tasks such as image recognition, action classification, and even autonomous driving \cite{paul2022}. The success of ViT has recently been attributed to their ability to process visual scenes like humans. This is particularly evident in their broader receptive fields compared to other model architectures and the distinct patterns of errors they exhibit \cite{tuli2021}. Yet, utilizing ViT in real-world situations like driving poses challenges stemming from their limited interpretability and the absence of frameworks for direct human guidance.

We propose a novel approach \footnote[1]{Code and data available at: \href{https://github.com/schko/fixatt}{github.com/schko/fixatt}} to tackle these challenges: incorporating eye-tracking data into ViT. Eye fixations offer a reliable measure of visual behavior and are often used to analyze human perception of intricate scenes \cite{cavanagh2014}. Moreover, the attention mechanism intrinsic to ViT has been leveraged to study their interpretation of images and videos \cite{gildenblat2020, bertasius2021}. We conduct experiments on two datasets involving human turn-taking decisions in virtual and real-world scenarios. Our primary objective is to uncover the relationship between human fixations and model attention to enhance the precision and reliability of decisions achieved by their combination.

First, we highlight differences in decision-making between humans and transformers in driving scenarios (left and right turning choices) under various types of uncertainty, expressed as opacity or contrast of the visual scene. Humans tend to mitigate localized uncertainty by fixating on fewer scene regions for extended durations. We then propose the fixation-attention intersection (FAX) loss, which calculates the intersection (dot product) of the model’s attention map with the human fixation map. Experiments with both virtual reality and real-world datasets demonstrate that gaze integration enhances ViT accuracy in uncertain conditions (fog, bad weather/light conditions). When trained with FAX loss, the model’s attention aligns with human gaze patterns, showing the potential for predicting human gaze. Additionally, we release a novel dataset of human driving decisions collected in virtual reality to study turn behavior.



\section{Related work}
\label{sec:related_work}

The parallel between human and machine vision has attracted considerable interest. Recent approaches have emphasized transformers' self-attention attributes and receptive fields, which mirror the human visual system \cite{tuli2021}, are robust to occlusions and perturbations \cite{touvron_training_2021}, generalize to multiple problems, and highly accurate compared to convolutional networks \cite{naseer_intriguing_2021, geirhos_imagenet-trained_2022}. Transformer-based architectures have shown high accuracy in predicting various eye tracking measures, such as types of eye behavior, gaze paths, or saliency maps \cite{lou_transalnet_2022}. Previous research has primarily explored retrospective comparisons between human and model attention using transformer-based networks \cite{sood-etal-2020-interpreting, Mehrani2023SelfattentionIV}, or methods like knowledge distillation through teacher-student model designs. An alternative approach \cite{sood2023-multimodal-integration} in the context of Visual Question Answering (VQA) uses saliency-predicting models to guide model attention by integrating human-like attention across both image and text modalities. While these approaches provide valuable insights into aligning machine with human attention, they do not directly integrate human gaze into the model training process. Our proposed method combines the advantage of larger receptive fields of ViT with the ability of the human visual system to gather task-relevant information from complex scenes quickly \cite{renninger2007} during model training.

Eye-tracking in humans can provide many insights into the behavioral and neural dynamics that underlie the flexible decision-making required in tasks such as driving. Current research in driving behavior has focused on integrating eye tracking to driver monitoring systems to index the driver’s attention and alertness as given by fixation coordinates and pupil dilation \cite{cavanagh2014, palazzi_predicting_2018}. Recent gaze-driven driving research has also been focused on identifying when a driver is distracted or not paying attention to the road. These systems can detect inattention by analyzing gaze direction and duration and provide warnings or interventions to maintain safety \cite{ahlstrom2013, dorazio2007}. Eye-tracking can aid in predicting the driver's intentions, such as lane changes or turns \cite{degee2014}. By analyzing gaze-based indices, an automated system can anticipate the driver's maneuvers, adjust its behavior accordingly, and potentially use gaze patterns to ensure safety and efficiency. 

Humans are notably successful in performing sensorimotor decisions under uncertainty compared to their artificially intelligent counterparts \cite{barabas_current_2017, cardenas_few_2020, krakauer_human_2011}. In tasks such as making a right turn onto a street, humans can infer and integrate information across spatial, temporal, and sensory modes for optimal and efficient decisions \cite{gallivan_decision-making_2018}. Recent advances in state-of-the-art robotics aim to integrate templates of the processes that underlie sensorimotor decision-making in humans to improve existing flexibility in decision-making \cite{noda2014, peternel2017}. Prior studies have revealed that visual attention is a critical cognitive process in performing sensorimotor decision-making tasks in the information processing stage but not in the motor planning stage \cite{dorazio2007, schall1998}. Visual attention is often evaluated with eye tracking technologies as visual information available to a subject depends on the field of view and the position of the pupil \cite{renninger2007}.

\section{Proposed Methods} 
\label{sec:proposed_methods}

\subsection{Baseline Vision Transformer}
Following the original ViT architecture \cite{dosovitskiy2020}, the representative frame $\mathbf{x} \in \mathbb{R}^{H \times W \times C}$ from the premotor period prior to motor action is divided into $N$ non-overlapping patches of size $P \times P$, which are then flattened to form $\mathbf{x}_p \in \mathbb{R}^{N \times (P^2 \cdot C)}$, where $(H, W, C)$ are the dimensions of the input frame and $N = HW/P^2$ is the total number of patches. Given that $a, \, a = 1\ldots A$, represents the number of attention heads and $l, \, l = 1\ldots L,$ the number of layers in the ViT model, we choose to implement a ViT model with $L=12$ layers and $A=12$ attention heads. The weights for each attention head across layers are given by:

\begin{equation}\label{self-attention}
    \mathcal{A}_{(l, a)} = \mathrm{softmax} \left( \frac{\mathbf{q}_{(l, a)}\mathbf{k}^T_{(l, a)}}{\sqrt{D_h}}\right)
\end{equation}

where $\mathcal{A}_{(l, a)} \in \mathbb{R}^{(N+1) \times (N+1)}$ and $D_h = D / A$, where $D$ is the embedding size. We convert the attention weight matrix $\mathcal{A}_{(l, a)}$ into a vector $\mathbf{a}_{(l, a)} \in \mathbb{R}^{N}$ by averaging over patches, while excluding the CLS token. The resulting vector $\mathbf{a}_{(l, a)}$ illustrates how the model, at head $a$ and layer $l$, assigns attention to different image patches and is used to visualize attention maps.

In our specific application, which centers around predicting left or right turns based on the premotor period frame, we employ binary cross entropy loss ($\mathcal{L}_{BCE}$) as the loss function for the baseline ViT model.

\begin{equation} \label{BCE}
    \mathcal{L}_{BCE} = - c_{1} \cdot \mathrm{log}(m_{1}) -c_{2} \cdot \mathrm{log}(m_{2})
\end{equation}

where $c_{1}, \, c_{2} \in \{0, 1\}$ denotes the two classes (left, right) and $m_{1}, \, m_{2} \in [0, 1]$ represent the predicted probabilities for the left and right class respectively.


\subsection{Fixation Maps}
Fixation maps $\mathbf{f} \in \mathbb{R}^{H \times W}$ represent the aggregate eye gaze during the premotor period and match the size of input frames. We define flattened patches of fixation map $\mathbf{f}_p \in \mathbb{R}^{N \times (P^2 \cdot C)}$, similar to the approach in the baseline ViT model. Additionally, we resize and flatten the original fixation map $\mathbf{f}$ to produce the vector $\mathbf{f}_{red} \in \mathbb{R}^{N}$, which has same size as the vector $\mathbf{a}_{(l, a)}$ of the ViT model and is used to compute the similarity between the attention and fixation maps.

\subsection{Fixation-Attention Intersection (FAX) Loss}
To better guide the baseline ViT model to simulate human attention, we introduce a novel fixation-attention intersection loss $\mathcal{L}_{FAX}$ to improve the model's ability to capture human-like attention patterns during training. This loss quantifies the average intersection ($\mathcal{I}$) as the dot product between ViT attention weights $\mathbf{a}_{(l, a)}$ of all heads across all layers and the reduced human fixation map $\mathbf{f}_{red}$. To ensure compatibility between $\mathcal{L}_{INT}$ and $\mathcal{L}_{BCE}$, we apply a sigmoid function to the average intersection $\mathcal{I}$. This prevents $\mathcal{I}$ from being too small, which could otherwise make $\mathcal{L}_{INT}$ disproportionately large and dominate $\mathcal{L}_{BCE}$. Restricting $\mathcal{I}$ to a range between 0.5 and 1 keeps both loss terms on a comparable scale, allowing their effective combination.

\begin{equation} \label{dot_product_intersection}
    \mathcal{I} = \frac{\sum\limits_l^L \sum\limits_a^A \mathbf{a}_{(l, a)} \cdot \mathbf{f}_{red}}{L \cdot A}
\end{equation}

\begin{equation} \label{intersection_loss}
     \mathcal{L}_{INT} = \frac{1}{\mathrm{sigmoid}(\mathcal{I})}
\end{equation}

Finally, we define $\mathcal{L}_{FAX}$ by combining $\mathcal{L}_{INT}$ with the original classification loss $ \mathcal{L}_{BCE}$:

\begin{equation} \label{custom_loss}
     \mathcal{L}_{FAX} = (1 - \lambda) \cdot  \mathcal{L}_{BCE} + \lambda \cdot  \mathcal{L}_{INT} ,
\end{equation}

where $\lambda$, $\lambda \in [0, 1]$, is the hyperparameter used for the weighted addition of the two losses.  To determine the optimal value of $\lambda$ for our experiments, we systematically evaluated a range of $\lambda$ values, namely \{0.01, 0.1, 0.2, 0.8, 1\}.

\subsection{Peripheral Masking of the Input} 
\label{sec:peripheral_masking_description}
\begin{figure}[h]
    \centering
    \includegraphics[width=\columnwidth]{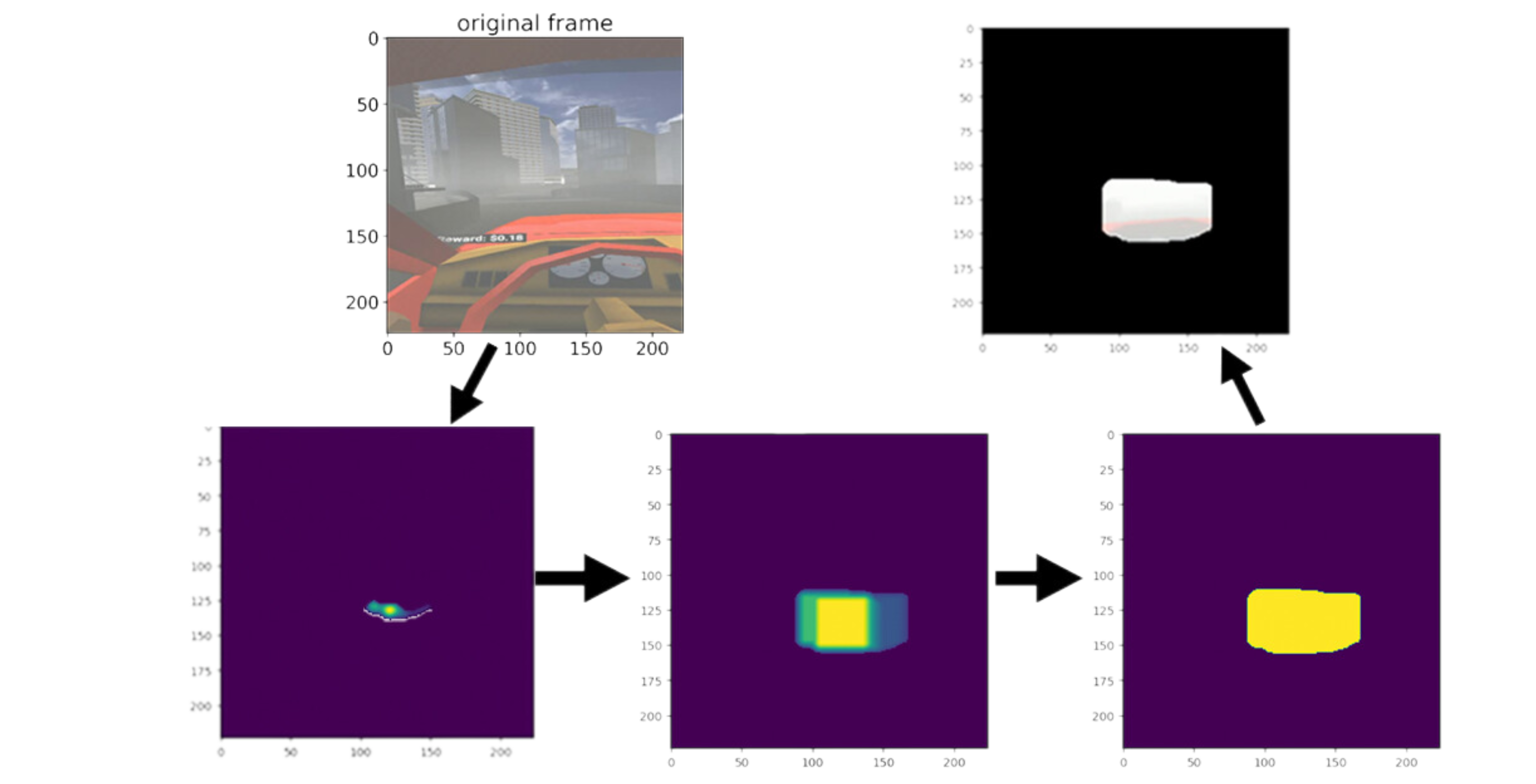}
    \caption{Peripheral masking of the input.}
    \label{fig:peripheral_masking}
\end{figure}

Peripheral masking involves the removal of regions outside the visual periphery within the frame. This is achieved by expanding the fixation area within fixation maps ($\mathbf{f}_p$) and zeroing all pixels outside of this area (Fig. \ref{fig:peripheral_masking}). To study the importance of human-fixated regions, we transform input data using peripheral masking for both datasets and compare the performance with random rotation and translation of the mask in a ``random masking'' control.

\section{Datasets}
\label{sec:datasets}

\begin{table*}[h]
    \caption{Overview of VR and DR(eye)VE datasets.}
    \centering
    \begin{tabular}{ccccccccc}
        \toprule
         Dataset & Uncertainty & Train Set & Valid Set & Test Set & Left Turns(\%) & Density & Contrast \\
        \midrule
        VR & High & 2015 & 356 & 599 & 48.9 & 0.65 & -  \\
        VR & Low & 2020 & 356 & 600 & 51.1 & 0.24 & -  \\
        DR(eye)VE & High & 236 & 55 & 73 & 51.0 & - & 0.11 \\
        DR(eye)VE & Low & 236 & 55 & 73 & 46.7 & - & 0.39  \\
        \bottomrule
    \end{tabular}
    \label{tab:results}
\end{table*}

\subsection{VR Dataset}
The VR dataset was collected as a part of a more extensive study on closed-loop brain-computer interface \cite{Koorathota_2023} (Fig.~\ref{fig:adadrive}). 10 participants were recruited to complete a boundary avoidance task (BAT), presented by the HTC Vive Pro Eye VR headset, in a virtual city environment with varying visual noise opacity. Using the Logitech G steering wheel, participants were instructed to drive a simulated car toward target locations. Steering wheel data and a video of the driving scene were recorded throughout the driving sessions. We identify motor actions through a simple peak and trough-detection technique on the steering wheel channel. Using a non-overlapping, look-behind window of 750ms, we assured that the peak we encountered was the true peak in steering wheel activity. The transformer models in our study were trained with frames corresponding to individual left-right turn motor actions. There are 6006 frames identified to be associated with a left or right turn, with 3293 left-turn frames. In addition to the video and steering information, eye-tracking data was collected using an HTC VIVE Pro Eye headset. Gaze coordinates from the eye tracker was used to construct the fixation map for each input frame. This map was computed by aggregating fixation data spanning a 3-second duration of the premotor period. We use the last frame in this premotor period as the input to all models.

\subsection{DR(eye)VE Dataset}

DR(eye)VE \cite{alletto_dreyeve_2016} is a publicly available driving dataset collected in real-world conditions across various landscapes, weather conditions, and times of day. The dataset contains gaze coordinates, driving speed, and course information for more than 500,000 frames. Geo-referenced locations are also available approximately every 25 frames. Because steering wheel data was unavailable, we used a combination of relative car positions, global positioning coordinates, and driving speed to identify left and right turn actions. To ensure the accuracy of our automated turn detection pipeline, at least two of the authors reviewed the videos manually and annotated frames corresponding to left or right turns. After review and validation of video frames, we identified 728 frames associated with a left or right turn, with 348 left-turn frames. Eye tracking data was collected using an SMI ETG 2w sensor. Fixation maps were computed using the same method for the VR dataset with a premotor period of 1 second. We use the first frame in this premotor period as the input to all models, driven by the relatively narrow field of view of the DR(eye)VE scene camera compared to the VR scene camera.

\subsection{Uncertainty in Visual Scene}
In our study, uncertainty refers to the visual conditions within a scene that hinder a driver's ability to clearly recognize objects. These conditions include various factors such as fog, weather, and lighting, which differ between the VR and real-world driving datasets. Although the specific definition of uncertainty varies between these datasets, the common factor is that high uncertainty corresponds to scenes where the driver has difficulty to perceive objects clearly, whether due to dense fog in the VR environment or adverse weather and lighting conditions in the real-world dataset.

In the VR dataset, we manipulated visual uncertainty on a trial-by-trial basis by adjusting the visual noise opacity parameter, as described in \cite{Koorathota_2023}. This adjustment simulates the type of white, 1/f noise typically found in visual search tasks, and participants perceived it through the varying density of fog in the virtual city environment. High noise opacity represented higher uncertainty, making it more difficult for participants to distinguish objects and navigate the scene.

In the DR(eye)VE dataset, which represents real-world driving scenarios, uncertainty was quantified by computing the average contrast across the entire image. This contrast was calculated by averaging the local pixel contrast, determined from the minimum and maximum luminance within a $5 \times 5$ kernel around each pixel. Lower average contrast indicates higher uncertainty, as it reflects poorer visibility conditions, such as bad weather or low-light situations, where distinguishing objects becomes more challenging.

We summarize the two datasets in Table~\ref{tab:results}. We also provide a video of an example turn from both datasets (see supplementary material).




\begin{figure*}[h]
    \centering
    \includegraphics[width=\textwidth]{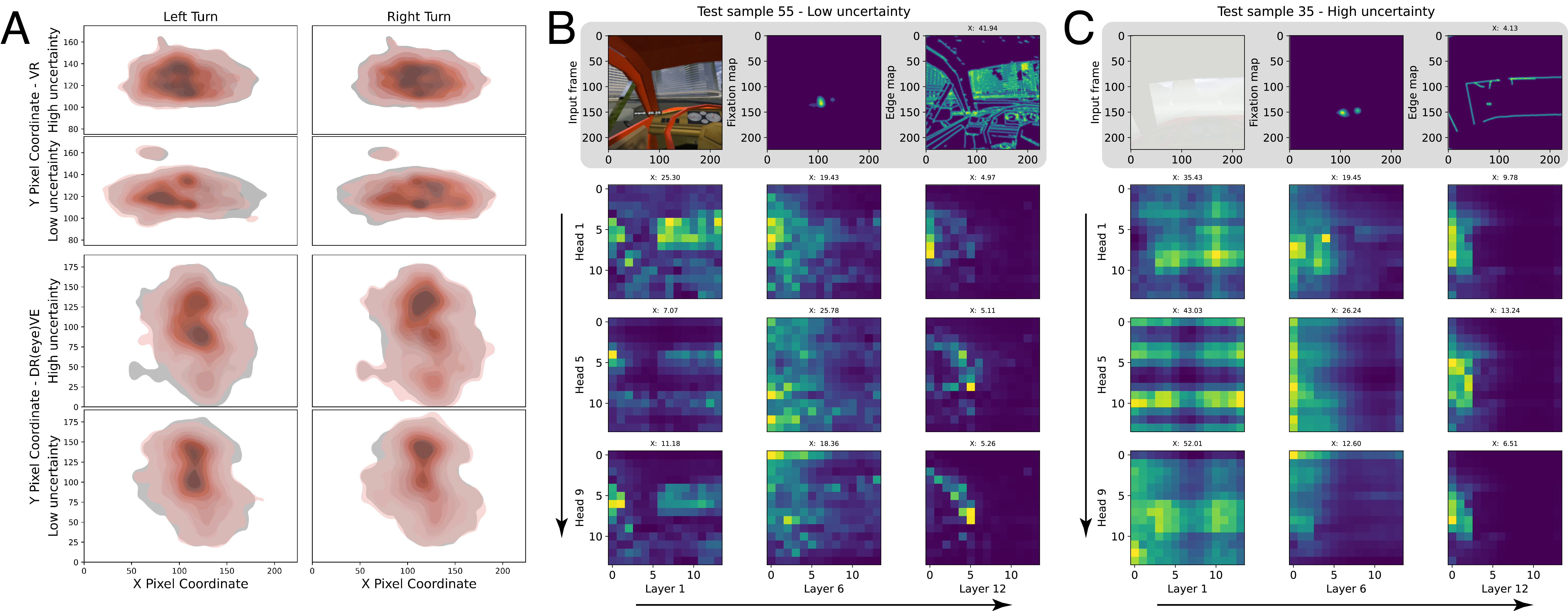}
    \caption{(A) KDE plot illustrating the distribution of fixations across pixel coordinates (x and y) across all test sample frames in the VR and DR(eye)VE datasets. Fixations are extracted from and aggregated over the premotor period prior to motor decisions. Higher density distribution indicates higher fixation duration. Class-specific (left or right) distributions are denoted in red; the overall distribution is gray. (B and C) Qualitative ViT results from two test samples corresponding to low (B) and high (C) uncertainty conditions in the VR dataset. X = dot product similarity between fixation and respective activation map. Only weights from 3 heads across 3 layers, corresponding to the first, middle, and last layers, respectively, are shown.}
    \label{fig:qualitative_uncertainty}
\end{figure*}

\section{Results}
\label{sec:results}

\subsection{Comparing Human and Model Attention Under Uncertainty}
\label{sec:human_vs_model_attention}

To establish the importance of integrating eye gaze into ViT, we analyze the complementary nature of human and model attention under uncertainty. This is demonstrated through a comparison of human fixation maps (Fig.~\ref{fig:qualitative_uncertainty}A) with the attention weights from the 12-layer baseline ViT model (Fig.~\ref{fig:qualitative_uncertainty}B, C) in the VR dataset, selected for its larger sample size.

The distribution of fixations in both datasets (gray) indicates that viewing time is concentrated around the center of the frames, although relatively sparse, with a larger fixation area for DR(eye)VE than the VR dataset. This difference is likely because participants were goal-directed in the VR dataset, focusing on avoiding the boundary and navigating through the environment with different levels of visibility. In comparison, in the DR(eye)VE dataset, participants needed to remain vigilant about other vehicles, attend to road signs, and ensure a safe driving experience by exploring the scene for unexpected events. The high uncertainty conditions across both datasets result in longer relative fixation durations in fewer regions. This finding is consistent with existing literature \cite{renninger2007} and suggests that humans minimize local spatial uncertainty in low-visibility scenarios through longer fixation time in fewer areas rather than minimize global uncertainty through an exploration strategy.

During the decision-making process concerning left or right turns, the attention weights (as depicted in Fig.~\ref{fig:qualitative_uncertainty}B, C) exhibit broader scene coverage than human fixation maps. This phenomenon is characterized by increased attention across the entire frame, particularly in high-uncertainty scenarios. Furthermore, the overall attention across ViT layers differs with depth. In shallower layers, attention is dispersed, primarily capturing edge-related scene details, while in deeper layers, attention is more concentrated, integrating contextual information. Our results suggest that transformer models may seek to minimize global over local uncertainty, the opposite of human strategy. Thus, human gaze may provide the model with information on which regions of the frame may be more relevant for resolving uncertainty, allowing more accurate learning.

\subsection{Layer Pruning in Vision Transformers Based on Similarity to Human Attention}
\label{sec:layer_pruning}

\begin{figure}
    \centering
    \includegraphics[width=\columnwidth]{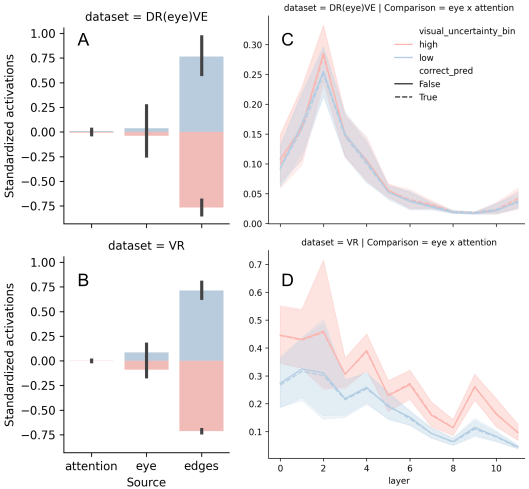}
    \caption{(A, B) Total, standardized sum of activations, by uncertainty split, for both datasets. We define total activation in the baseline ViT as the sum of attention weights across layers and heads. Total fixation refers to the pixel-wise sum of fixation maps, a measure of the overall fixation area. Total edge activation refers to the pixel-wise sum of edge maps. (C, D) The similarity between attention weights across layers and fixation maps, using Eq.~\ref{dot_product_intersection}. Results are aggregated from all test samples on the best-performing, 12-layer baseline ViT. Line color shows the uncertainty split of the test samples, while line style shows whether ViT classified the motor action correctly. Error band shows the 95\% CI. }
    \label{fig:aggregate_results}
\end{figure}

To understand how model attention overlaps with edges and fixations from the scene, we measure overall attention and compute the similarity between fixation maps and layer-specific attention maps using the dot product described in Eq.~\ref{dot_product_intersection}.  First, the total ViT attention is not notably different between low and high uncertainty test samples. In contrast, the total number of fixations and edges varies by uncertainty condition (Fig.~\ref{fig:aggregate_results}A, B). In other words, high visibility (i.e., low uncertainty) results in a larger fixation area and more detected edges. Still, the model does not employ a different strategy in its attention overall. This suggests that the model may find fixation data beneficial for parsing uncertainty through better distributing its attention weights spatially in the scene.

Following the observation that total model attention does not differ by visual uncertainty, we explored layer-specific attention and its overlap with human fixation maps, as shown in Figures~\ref{fig:aggregate_results}C, D. Observing a decline in this overlap beyond the fifth layer,  we pruned the ViT model to its first five layers, creating a 5-layer model (5-ViT). We also tested a single-layer ViT (1-ViT), as a control.

For both DR(eye)VE and VR dataset, the accuracy of the 5-ViT model is not significantly different than 12-ViT (Table \ref{tab:pairwise_comparison}), suggesting that reducing the model's complexity by half does not impact accuracy. However, reducing the model to a single layer (1-ViT) significantly lowers performance compared to 12-ViT across both datasets. This outcome demonstrates that models pruned to retain layers with the highest alignment to human attention can maintain or even enhance performance, especially in scenarios with smaller datasets like DR(eye)VE, where the 5-ViT model shows an improvement, though not statistically significant, over the 12-ViT model.

\subsection{Assessing the Impact of Human Eye Gaze on Task Performance}
\label{sec:impact_of_gaze_on_task_performance}
Here, we investigate the contribution of human eye gaze information to the task of predicting left or right turns. In particular, we evaluate whether gaze data during the premotor period suffices for accurately predicting a left or right turn. Our experimental setup includes a baseline model, a 12-layer Vision Transformer (12-ViT), processing the entire image frame. This model's performance was benchmarked against two approaches: peripheral masking (Section \ref{sec:peripheral_masking_description}) and a dummy classifier that predicts the turn direction based on the assumption that drivers' gaze towards the right or left side of the frame indicates the corresponding turn direction (see Appendix).

Considering the task's reliance on spatial scene information, we evaluated the efficacy of a dummy classifier in leveraging eye gaze data for turn prediction. The performance of this classifier established a benchmark for evaluating our proposed methods, achieving 50.48 $\pm$ 4.19\% accuracy in the DR(eye)VE dataset and 64.67 $\pm$ 1.33\% in the VR dataset (Table~\ref{tab:top_5_accuracy}). These results indicate that gaze information alone can accurately predict driving decisions in the VR dataset for a significant proportion of cases. The effectiveness in the VR context is attributed to its streamlined design, featuring roads without complex navigational choices or directional changes, where drivers' turn directions align with their gaze. Conversely, the real-world driving environment presents complex scenarios with external factors, such as pedestrians and other vehicles, thereby complicating the direct correlation between gaze direction and turning decisions observed in the VR environment.

We also used peripheral masking to assess whether focusing solely on the driver's fixation area within the scene yields enough information for accurate task completion. Across both datasets, our analysis revealed no significant difference in total accuracy between the full-frame baseline model (12-ViT) and the model trained on peripherally masked frames (Peripheral-ViT), accounting for both high and low uncertainty conditions (Table \ref{tab:pairwise_comparison}). These results suggest that while the fixation area is critical for classification, the surrounding scene holds valuable information as well. Consequently, we opted for the FAX loss approach, which rewards the model for focusing more on the fixation area without neglecting the surrounding scene information.

\begin{figure}[h]
    \centering
    \includegraphics[width=\columnwidth]{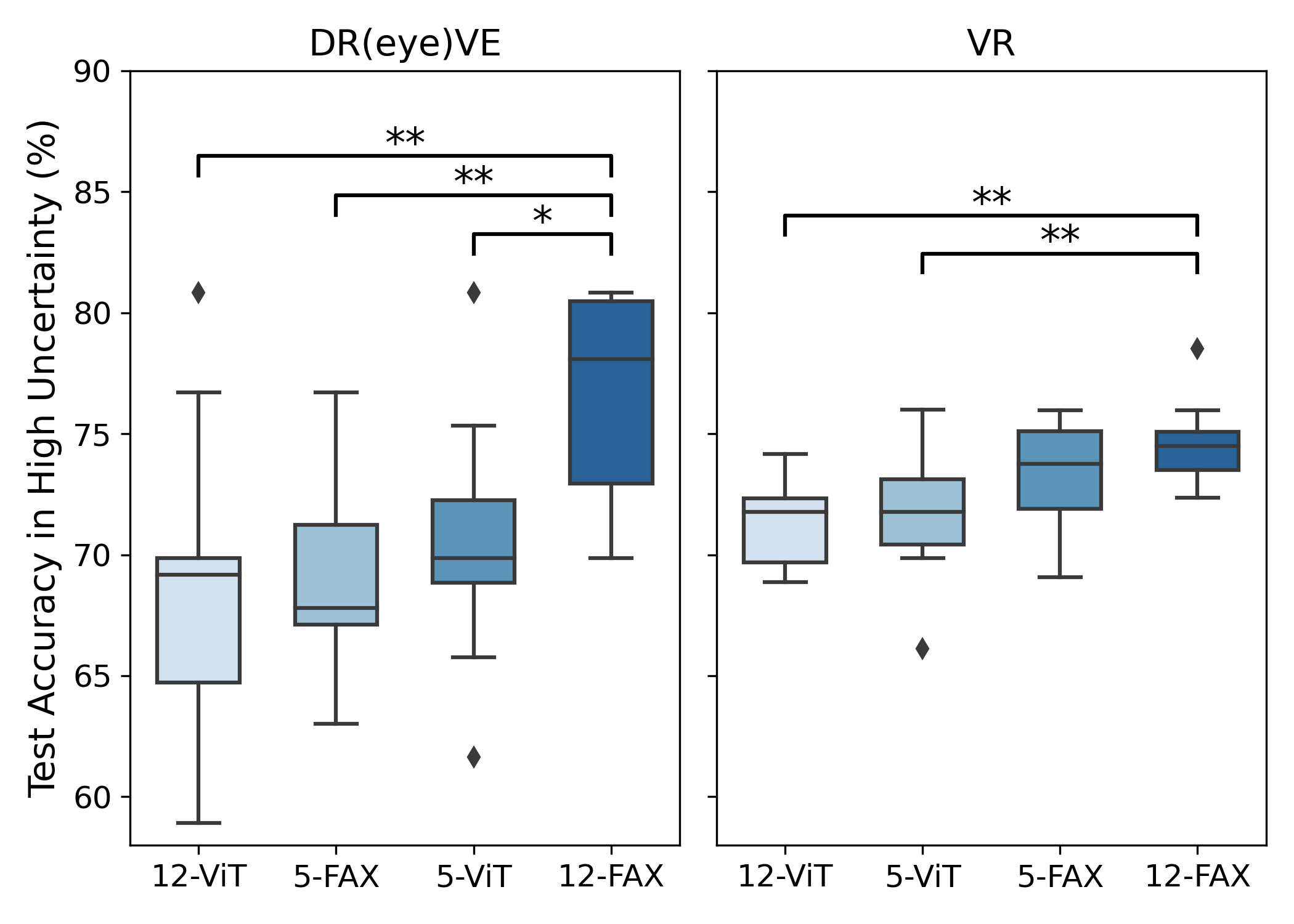}
    \caption{Boxplots displaying the test accuracy in high uncertainty of the top performing models on the DR(eye)VE and VR datasets. 12-ViT and 5-ViT denote Vision Transformer models with 12 and 5 layers, respectively; 12-FAX and 5-FAX represent equivalent ViT models trained with the FAX loss, with the optimal $\lambda$ value in each case. The Mann-Whitney U test assesses statistical significance (* p < 0.05, ** p < 0.01, *** p < 0.001).}
    \label{fig:accuracy}
\end{figure}

\begin{table}[h]
\centering
\caption{Top performing models based on test accuracy (mean $\pm$ std). Models are compared against the baseline accuracy of the dummy classifier.} 
\begin{tabular}{@{}ccccc@{}}
\toprule
& Model & High Uncertainty & Low Uncertainty & Total \\
\midrule
\multirow{6}{*}{\rotatebox{90}{DR(eye)VE}} & \textit{Dummy} &  49.73 $\pm$ 8.12  &  51.23 $\pm$ 4.80  &  50.48 $\pm$ 4.19  \\
& 12-ViT & 68.77 $\pm$ 6.42 & 64.11 $\pm$ 7.36 & 66.44 $\pm$ 6.19 \\
& 5-FAX & 69.18 $\pm$ 4.44 & 70.82 $\pm$ 4.96 & 70.00 $\pm$ 4.14 \\
& 5-ViT & 70.55 $\pm$ 5.18 & 69.45 $\pm$ 6.68 & 70.00 $\pm$ 5.35 \\
& 12-FAX & 76.58 $\pm$ 4.26 & 71.23 $\pm$ 4.03 & 73.90 $\pm$ 3.39 \\
\midrule
\multirow{6}{*}{\rotatebox{90}{VR}} & \textit{Dummy} &  71.15 $\pm$ 1.66 &  58.38 $\pm$ 2.03  &  64.67 $\pm$ 1.33  \\
& 12-ViT & 69.75 $\pm$ 6.50 & 60.47 $\pm$ 4.67 & 65.13 $\pm$ 5.34 \\
& 5-ViT & 71.68 $\pm$ 2.67 & 61.89 $\pm$ 2.47 & 66.71 $\pm$ 1.71 \\
& 5-FAX & 73.38 $\pm$ 2.25 & 61.24 $\pm$ 1.70 & 67.31 $\pm$ 1.22 \\
& 12-FAX & 74.66 $\pm$ 1.72 & 61.25 $\pm$ 1.97 & 67.88 $\pm$ 1.31 \\
\bottomrule
\end{tabular}
\label{tab:top_5_accuracy}
\end{table}

\subsection{Gaze Integration Improves ViT Accuracy in Uncertain Conditions}
\label{sec:gaze_integration_improves_vit_accuracy}

\begin{table*}[ht]
\centering
\caption{Pairwise Mann-Whitney U Test Comparisons.}
\begin{tabular}{@{}cccccccccc@{}}
\toprule
&  & \multicolumn{4}{c}{DR(eye)VE} & \multicolumn{4}{c}{VR} \\
\cmidrule(lr){3-6} \cmidrule(lr){7-10}
Model 1 & Model 2 & \multicolumn{2}{c}{Total} & \multicolumn{2}{c}{High Uncertainty} & \multicolumn{2}{c}{Total} & \multicolumn{2}{c}{High Uncertainty}  \\
\cmidrule(lr){3-4} \cmidrule(lr){5-6} \cmidrule(lr){7-8} \cmidrule(lr){9-10}
 & & p-value & reject & p-value & reject & p-value & reject & p-value & reject \\
\midrule
12-FAX & 12-ViT & 0.005 & True  & 0.007 & True  & 0.064 & False & 0.002 & True \\
12-FAX & 5-ViT  & 0.110 & False & 0.016 & True  & 0.161 & False & 0.007 & True \\
12-FAX & 5-FAX  & 0.044 & True  & 0.003 & True  & 0.405 & False & 0.326 & False \\
5-ViT  & 5-FAX  & 0.970 & False & 0.518 & False & 0.677 & False & 0.173 & False \\
12-ViT & 5-ViT & 0.161 & False  & 0.378 & False  & 0.570 & False & 0.850 & False \\
12-ViT & 1-ViT & 0.008 & True  & 0.002 & True  & 0.003 & True & 0.003 & True \\
12-ViT & Periph.-ViT & 0.000 & True  & 0.000 & True  & 0.006 & True & 0.140 & False \\
Dummy & Periph.-ViT & 0.677 & False  & 0.621 & False  & 0.058 & False & 0.096 & False \\
\bottomrule
\end{tabular}
\label{tab:pairwise_comparison}
\end{table*}

Next, we validate our findings from \ref{sec:human_vs_model_attention} that human eye gaze can resolve uncertainty. To do so, we trained both the full-layer and the ablated ViT models using the FAX loss to drive the model's attention towards areas aligned with human gaze. For both datasets, the top performing models include: 12-FAX, 5-FAX, 5-ViT, 12-ViT (Fig. \ref{fig:accuracy}). In the DR(eye)VE dataset, the 12-FAX model demonstrated a significant accuracy improvement over the 12-ViT model, with a mean increase of 7.46\%. Furthermore, the 12-FAX model exhibited a distinct, though not statistically significant, accuracy improvement over the 5-ViT model and a significant improvement over the 5-FAX model. Notably, in high uncertainty conditions, the 12-FAX model consistently outperformed the 12-ViT and 5-ViT models, underscoring the benefits of gaze integration. These results are consistent with our previous observations that human eye gaze contributes to resolving uncertainty and enhances ViT performance, especially in high uncertainty conditions. 

In the VR dataset, performance differences among models were not statistically significant in low uncertainty conditions. However, in high uncertainty scenes, the 12-FAX model significantly outperformed both the 12-ViT and 5-ViT models, further highlighting the importance of human gaze in resolving uncertainty. Additionally, there is no significant difference in accuracy between 5-FAX and 5-ViT across both datasets and all levels of uncertainty (Table \ref{tab:pairwise_comparison}), suggesting that shallower networks do not benefit as significantly as the full-layer model from the integration of human gaze data. Overall, our results support the conclusion from \ref{sec:human_vs_model_attention} that human gaze enhances the model’s attention, thereby improving ViT model performance in uncertain conditions.

\begin{figure*}
    \centering
    \includegraphics[width=\textwidth]{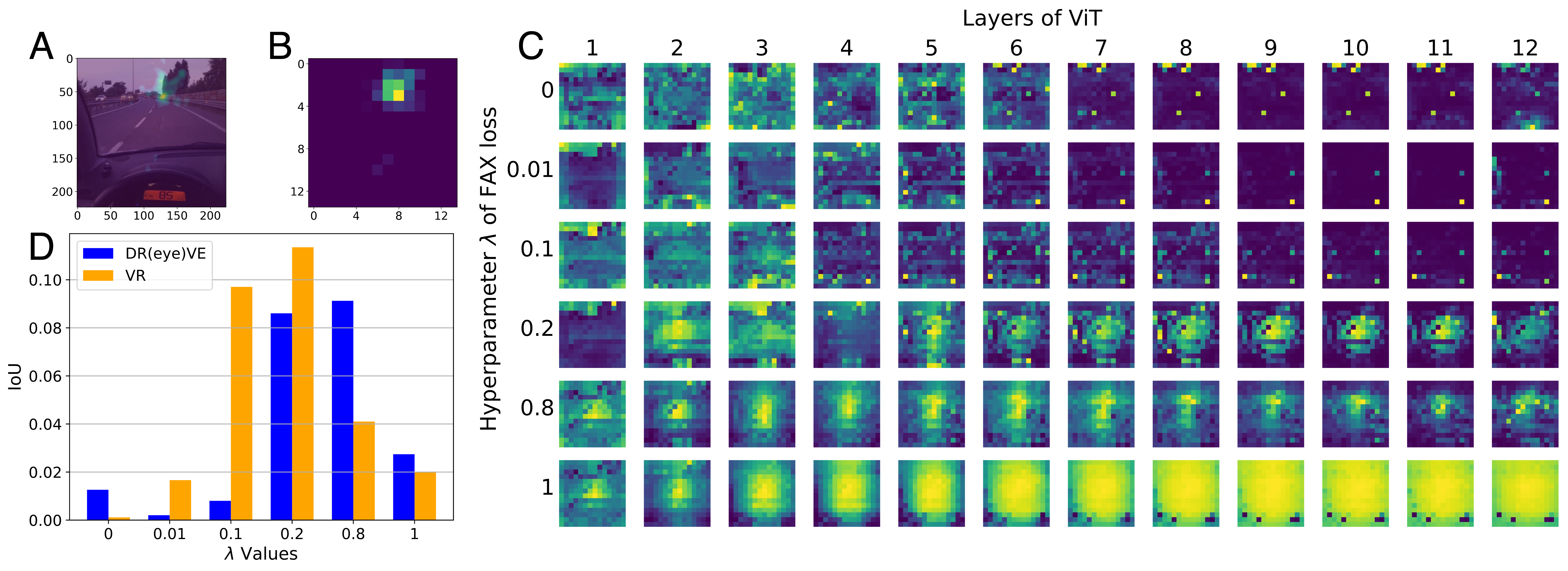}
    \caption{Training with FAX loss aligns model attention with human gaze. This figure shows the impact of FAX loss on the ViT model's attention maps for test set samples. (A) Input frame with overlaid human fixation map. (B) Human fixation data aligned to ViT attention map dimensions. (C) Average attention maps across all heads for each ViT layer, for distinct $\lambda$ values in FAX loss (Eq. \ref{custom_loss}), showing increasing resemblance to human fixation patterns with higher $\lambda$ values. (D) Intersection over Union (IoU) metric between attention and fixation maps for the test set samples for all $\lambda$ values, quantifying alignment. These results demonstrate that optimal $\lambda$ values in FAX loss (e.g., $\lambda = 0.2, 0.8$ for DR(eye)VE and $\lambda = 0.1, 0.2$ for VR) lead to attention maps better resembling human fixation area, indicating the model's ability to predict human gaze in driving scenarios.}
    \label{fig:lambda_layers}
\end{figure*}

\subsection{FAX Loss Aligns Model Attention with Human Gaze}

Lastly, we present qualitative results displaying the average attention map across all heads for each layer of the ViT model with varying $\lambda$ values, from $\lambda = 0$ (where only $\mathcal{L}_{BCE}$ is used) to $\lambda = 1$ (where only $\mathcal{L}_{INT}$ is used), as illustrated in Fig.~\ref{fig:lambda_layers}. An increase in $\lambda$ (from left to right in the figure) correlates with attention maps increasingly resembling human fixation patterns, except at $\lambda = 1$, where the model disperses attention across the frame to mimic human fixations. In the earlier layers, attention is distributed globally across the image, while subsequent layers progressively shift to more precise attention regions correlating with human fixations, especially at optimal $\lambda$ settings, such as 0.2 or 0.8. Notably, at $\lambda = 0.8$, which aligns with the highest accuracy in the DR(eye)VE dataset, there is the highest alignment between attention maps and human fixation data. To quantitatively support these observations, we calculated the Intersection over Union (IoU) metric between attention and human fixation maps for all $\lambda$ values (Fig.~\ref{fig:lambda_layers} D). The IoU was calculated by converting heatmaps to binary masks using a threshold of 0.4, computing the intersection and union of these masks, and averaging the IoU across all heads and layers for all test set samples of the two datasets.

The results demonstrate that ViT, when trained with FAX loss at an optimal $\lambda$ value, can adopt an attention strategy similar to humans. By integrating gaze, ViT learn to focus on specific regions, reducing local uncertainty without compromising their broader receptive field. This significantly enhances model performance in driving contexts and could also apply to other areas where transformers are employed for image and video tasks.

\section{Limitations and Future Work}
While our study demonstrates the potential of integrating human gaze into Vision Transformers (ViT) for driving decision-making under uncertainty, several limitations warrant further investigation. Currently, our approach predicts turning decisions based on single frames. However, driving is inherently a dynamic task that requires processing sequences of frames to understand the broader context. Expanding our model to handle dynamic video data could enhance its applicability and performance in real-world scenarios. Future research should explore incorporating temporal information to better align the model with the sequential nature of human decision-making.

Moreover, while this work focuses on driving, the integration of human gaze has broader implications. Future studies should explore applying this approach in other domains where gaze data is valuable, such as medical imaging. For instance, training models on datasets that include radiologists' gaze data could enhance diagnostic accuracy and decision-making in clinical contexts \cite{Bulat-review-radiologistGaze-2024}.

\section{Conclusion}
\label{sec:conclusion}
Our paper establishes the critical role of human eye gaze in enhancing Vision Transformer (ViT) models for driving decision-making under uncertainty. A systematic comparison between human and model attention characteristics provided the groundwork for this integration. We introduced the fixation-attention intersection (FAX) loss, a novel methodology for incorporating gaze data into ViT. Notably, the application of FAX loss led to a significant performance improvement in ViT, particularly under high uncertainty conditions. These advancements demonstrate the potential of human-guided transformer models to refine driver behavior analysis and improve human-vehicle interaction, with broader implications for human-centered AI systems and human-computer interaction.

\section{Acknowledgments}
The authors would like to thank Pawan Lapborisuth, Josef Faller, and Ziheng Li  for their valuable contributions and support during the course of this study. The study was funded by a Vannevar Bush Faculty Fellowship from the U.S. Department of Defense (N00014-20-1-2027), a Cooperative Agreement with the Army Research Laboratory (W911NF-23-2-0067), and a Center of Excellence grant from the Air Force Office of Scientific Research (FA9550-22-1-0337). 

\bibliographystyle{ACM-Reference-Format}
\bibliography{sample-base}

\appendix
\section{Appendix}
\subsection{Comparative Performance of Models}
Comparative performance of all Vision Transformer (ViT) model variants on the VR dataset  and on the DR(eye)VE dataset under both low and high uncertainty combined (Table \ref{tab:combined_performance_table}). Metrics include test accuracy (\%), Area Under the Receiver Operating Characteristic Curve (AUC), and F1-Score. Results are presented as mean $\pm$ standard deviation computed from 10 independent training runs with distinct data splits. Model variants include 1-, 5-, and 12-layer ViT (denoted as 1-ViT, 5-ViT and 12-ViT) trained on image data using binary cross entropy (BCE) loss, and equivalent models trained with FAX loss (denoted as 1-FAX, 5-FAX and 12-FAX). Peripheral-ViT and Random Peripheral-ViT denote the baseline 12-layer ViT trained with ablated input, as outlined in the 'Proposed Methods' section. For the models trained with FAX loss, eye gaze was not used during inference. Instead, we loaded the trained weights to the vanilla ViT architecture and proceeded with predictions on the test data.

\begin{figure}[htbp]
    \centering
    \includegraphics[width=\columnwidth]{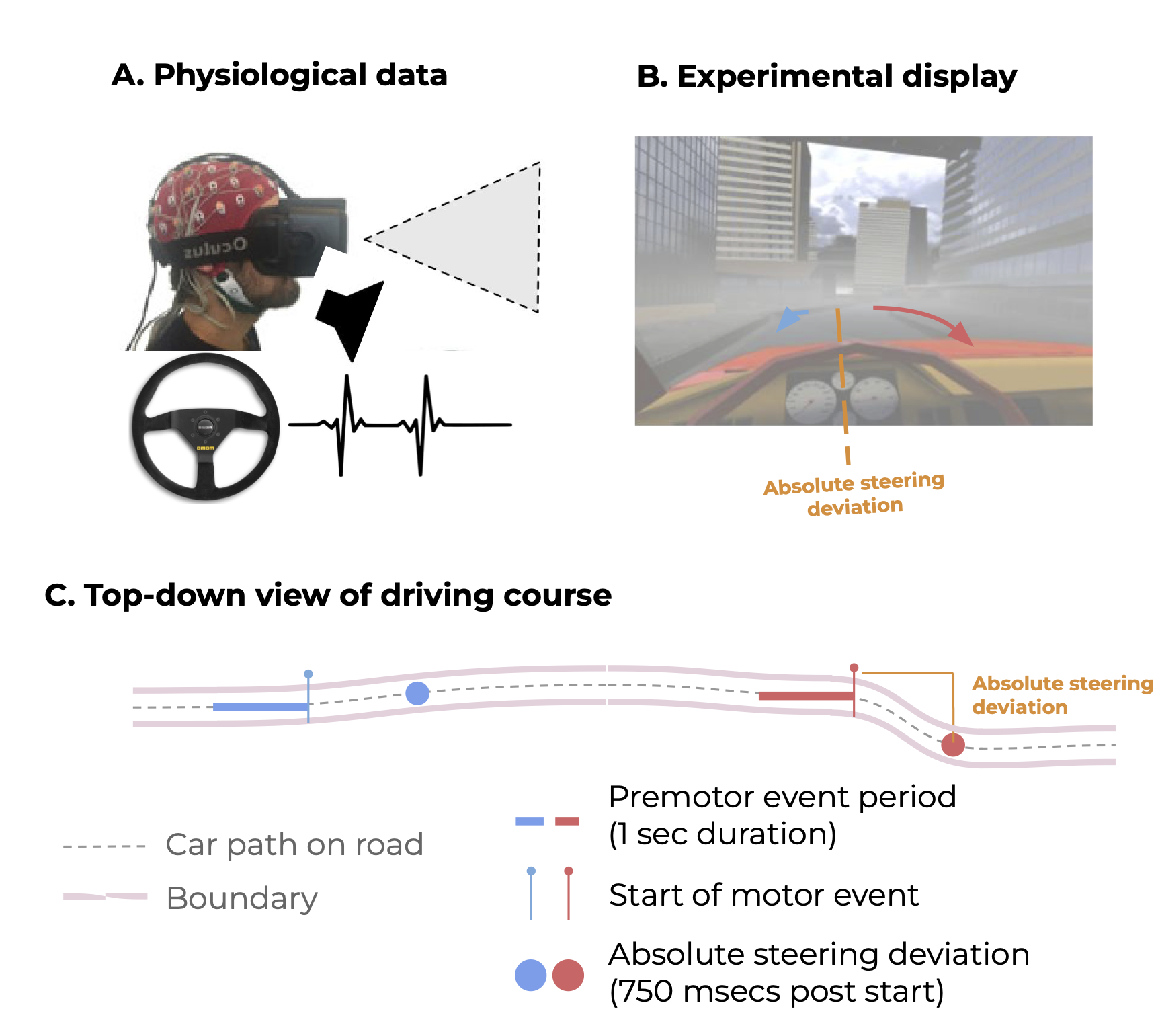}
    \caption{Adapted from \cite{Koorathota_2023}. Overview of how motor events were detected and premotor period is defined for the VR dataset. (A) We simultaneously collected neural data from EEG, autonomic measures using ECG, eye movements and pupil dynamics using a VR-headset embedded eye tracking system, and motor actions using a steering wheel. (B) Participants (n=10) performed 3 virtual reality driving task sessions, requiring boundary avoidance under time pressure and changing visual uncertainty. Their motor actions were recorded from the steering deviation as they were navigating a city environment. We analyzed direction-independent (i.e., absolute) steering deviation). Motor actions belong to a global trial with a set level of visual fog (opacity) in the environment that participants drove in. (C) The start of each motor action was marked using a peak detection method on the steering wheel data since this was most relevant to navigating the boundary avoidance task. The premotor periods of interest for this study were a fixed, 3-second interval before each event, and the intensity of the motor activity was determined by the post-event steer angle. Blue and red circles indicate events with low and high motor intensity, respectively.}
    \label{fig:adadrive}
\end{figure}

\subsection{Fixation Map and Edge Detection}
The fixation maps for both VR and DR(eye)VE datasets are generated by aggregating fixation data spanning a different duration of the premotor period. The fixation map for DR(eye)VE datasets was generated by utilizing the fixation coordinate for 10 frames, which is 0.4 seconds, whereas the VR dataset uses fixation data for 3 seconds. The reason for such a difference was, the fixation map for DR(eye)VE tends to have a larger fixation area due to the number of uncertain factors to consider. A shorter duration in this case generates a fixation map with a reasonable size.

The fixation map is generated based on the dimension of the original frame. The gaze coordinates for the VR, in particular, need to be re-scaled to match the frame dimension before the fixation map gets generated. The generation of the fixation map starts with computing a 2D Gaussian matrix with pre-defined variables. These variables are being used to further re-scale the gaze coordinate so that the coordinate aligns with the Gaussian matrix. The fixation map is generated by adding up the portion of the Gaussian matrix that corresponds to those gaze coordinates. There is a duration parameter available that can be used to adjust the weight of the gaze coordinate. The fixation map for the DR(eye)VE dataset is generated in the same fashion.

For edge detection, the original frame first goes through color conversion from color to grayscale. Afterward, Gaussian smoothing  is performed on the grayscale image with a 3x3 kernel. The purpose of this step is to improve the edge detection result. This blurred image finally feeds into the Canny edge detector with a lower threshold of 25 and an upper threshold of 50. The specific implementation details are provided in the \texttt{process\_driving\_video.py}.

\subsection{Peripheral Masking}
In our processing pipeline, we enhance the fixation heatmap to encompass a broader peripheral area, which we subsequently employ for generating masks. This augmentation of the heatmap accounts for the fact that peripheral regions can provide valuable insights into what is visually perceived around the central foveated point, as captured by eye tracking systems.

To achieve this, we perform a dilation operation on the fixation heatmap. Dilation involves expanding regions of high intensity, effectively enlarging the areas of interest. The dimensions of the kernel used for dilation are set empirically to 30x30 pixels within the context of a $224 \times 224$ input frame. This choice aligns with the approximate 30 degrees of visual angle that humans perceive in the mid-peripheral region. The specific implementation details are provided in the \texttt{create\_peripheral\_mask.py}.

\begin{table}
    \caption{Configuration of top performing models.}
    \centering
    \begin{tabular}{cccccc}
        \toprule
        \multirow{2}{*}{Model} & \multirow{2}{*}{Architecture} & \multirow{2}{*}{Layers} & \multirow{2}{*}{Loss} & \multicolumn{2}{c}{$\lambda$} \\
        \cmidrule(lr){5-6}
          &  &  &  & VR & DR(eye)VE  \\
         \midrule
         12-ViT & ViT & 12 & CrossEntr. & - & -\\
         5-ViT & ViT & 5  & CrossEntr. & - & -\\
         12-FAX & ViT & 12 & FAX & 0.1 & 0.8 \\
         5-FAX & ViT & 5 & FAX & 0.2 & 0.2 \\
        \bottomrule
    \end{tabular}
    \label{tab:results}
\end{table}

\begin{table*}
\caption{Performance metrics of ViT variants on the VR and DR(eye)VE datasets under both low and high uncertainty combined, showing test accuracy (\%), AUC, and F1-Score as mean $\pm$ standard deviation from 10 training runs with distinct data splits.}
\centering
\begin{tabular}{@{}>{\hspace{10pt}}c>{\hspace{10pt}}c>{\hspace{10pt}}c>{\hspace{10pt}}c>{\hspace{10pt}}c>{\hspace{10pt}}c>{\hspace{10pt}}c>{\hspace{10pt}}c>{\hspace{10pt}}c@{}}
\multicolumn{2}{c}{} & \multicolumn{3}{c}{VR Dataset} & \multicolumn{3}{c}{DR(eye)VE Dataset} \\
\cmidrule(lr){3-5} \cmidrule(lr){6-8}
Model & ${\lambda}$ & Accuracy (\%) & AUC & F1 & Accuracy (\%) & AUC & F1 \\
\midrule
\textit{Dummy} & - & 64.67 $\pm$ 1.33 &  0.65 $\pm$ 0.01 & 0.62 $\pm$ 0.02 & 50.48 $\pm$ 4.19 & 0.49 $\pm$ 0.04 & 0.60 $\pm$ 0.05 \\
\midrule
1-ViT & -  & 61.74 $\pm$ 1.55 & 0.62 $\pm$ 0.02 & 0.65 $\pm$ 0.02 & 58.22 $\pm$ 4.21 & 0.57 $\pm$ 0.05 & 0.62 $\pm$ 0.08 \\
5-ViT & -  & 66.71 $\pm$ 1.62 & 0.67 $\pm$ 0.01 & 0.65 $\pm$ 0.04 & 70.00 $\pm$ 5.08 & 0.70 $\pm$ 0.05 & 0.72 $\pm$ 0.07 \\
12-ViT & - & 65.13 $\pm$ 5.07 & 0.65 $\pm$ 0.05 & 0.61 $\pm$ 0.20 & 66.44 $\pm$ 5.87 & 0.66 $\pm$ 0.05 & 0.69 $\pm$ 0.08 \\
\midrule
12-FAX & 0.01 & 66.17 $\pm$ 1.09 & 0.66 $\pm$ 0.01 & 0.67 $\pm$ 0.03 & 63.22 $\pm$ 6.03 & 0.63 $\pm$ 0.06 & 0.65 $\pm$ 0.10 \\
>> & 0.1      & 67.88 $\pm$ 1.24 & 0.68 $\pm$ 0.01 & 0.70 $\pm$ 0.02 & 67.47 $\pm$ 4.84 & 0.66 $\pm$ 0.06 & 0.72 $\pm$ 0.05 \\
>> & 0.2      & 67.36 $\pm$ 2.00 & 0.67 $\pm$ 0.02 & 0.68 $\pm$ 0.04 & 64.66 $\pm$ 8.00 & 0.64 $\pm$ 0.08 & 0.67 $\pm$ 0.15 \\
>> & 0.8      & 65.80 $\pm$ 1.98 & 0.66 $\pm$ 0.02 & 0.68 $\pm$ 0.02 & 73.90 $\pm$ 3.22 & 0.74 $\pm$ 0.03 & 0.76 $\pm$ 0.05 \\
>> & 1        & 49.73 $\pm$ 1.53 & 0.50 $\pm$ 0.00 & 0.13 $\pm$ 0.27 & 46.44 $\pm$ 3.04 & 0.50 $\pm$ 0.00 & 0.26 $\pm$ 0.32 \\
\midrule
5-FAX & 0.01 & 66.11 $\pm$ 1.12 & 0.66 $\pm$ 0.01 & 0.66 $\pm$ 0.02 & 70.27 $\pm$ 4.04 & 0.70 $\pm$ 0.03 & 0.72 $\pm$ 0.07 \\
>> & 0.1     & 66.53 $\pm$ 1.58 & 0.67 $\pm$ 0.02 & 0.67 $\pm$ 0.02 & 69.04 $\pm$ 5.27 & 0.69 $\pm$ 0.05 & 0.69 $\pm$ 0.12 \\
>> & 0.2     & 67.31 $\pm$ 1.16 & 0.67 $\pm$ 0.01 & 0.67 $\pm$ 0.03 & 70.00 $\pm$ 3.93 & 0.70 $\pm$ 0.03 & 0.72 $\pm$ 0.06 \\
>> & 0.8     & 63.83 $\pm$ 2.56 & 0.64 $\pm$ 0.03 & 0.63 $\pm$ 0.08 & 66.51 $\pm$ 3.99 & 0.66 $\pm$ 0.04 & 0.68 $\pm$ 0.06 \\
>> & 1       & 49.83 $\pm$ 0.42 & 0.50 $\pm$ 0.00 & 0.25 $\pm$ 0.31 & 48.08 $\pm$ 4.53 & 0.50 $\pm$ 0.01 & 0.27 $\pm$ 0.33 \\
\midrule
1-FAX & 0.01 & 58.89 $\pm$ 4.55 & 0.59 $\pm$ 0.05 & 0.58 $\pm$ 0.15 & 62.05 $\pm$ 4.56 & 0.61 $\pm$ 0.04 & 0.67 $\pm$ 0.06 \\
>> & 0.1     & 51.63 $\pm$ 1.51 & 0.51 $\pm$ 0.02 & 0.54 $\pm$ 0.26 & 61.92 $\pm$ 4.41 & 0.61 $\pm$ 0.04 & 0.66 $\pm$ 0.07 \\
>> & 0.2     & 50.76 $\pm$ 0.81 & 0.51 $\pm$ 0.01 & 0.53 $\pm$ 0.26 & 61.16 $\pm$ 5.95 & 0.61 $\pm$ 0.04 & 0.60 $\pm$ 0.13 \\
>> & 0.8     & 50.42 $\pm$ 0.62 & 0.50 $\pm$ 0.00 & 0.53 $\pm$ 0.27 & 51.64 $\pm$ 4.18 & 0.51 $\pm$ 0.02 & 0.38 $\pm$ 0.28 \\
>> & 1       & 49.88 $\pm$ 0.29 & 0.50 $\pm$ 0.00 & 0.33 $\pm$ 0.33 & 46.85 $\pm$ 3.61 & 0.50 $\pm$ 0.01 & 0.17 $\pm$ 0.26 \\
\midrule
Peripheral-ViT & -        & 63.38 $\pm$ 1.30 & 0.63 $\pm$ 0.01 & 0.63 $\pm$ 0.05 & 51.23 $\pm$ 4.13 & 0.51 $\pm$ 0.04 & 0.54 $\pm$ 0.10 \\
Random Periph.-ViT & - & 52.90 $\pm$ 1.10 & 0.53 $\pm$ 0.01 & 0.53 $\pm$ 0.08 & 48.42 $\pm$ 4.01 & 0.48 $\pm$ 0.04 & 0.50 $\pm$ 0.06 \\
\bottomrule
\end{tabular}
\label{tab:combined_performance_table}
\end{table*}

\subsection{Dummy Classifier}
The dummy classifier is based on the simple assumption that a driver's gaze direction within an image frame -either right or left- indicates the intended turning direction. This classifier splits each image into two equal halves (left-right) and calculates the sum of pixel values for each side. A prediction for a right or left turn is then made based on which half of the image has a greater sum of pixel values, suggesting that a higher sum indicates the driver's focus area and, by extension, the turn direction. This approach establishes a baseline for comparison, illustrating the extent to which the spatial information inferred from gaze can address the task. Consequently, it allows for an evaluation of the enhancements brought by integrating gaze data into ViT models. The specific implementation details are provided in the \texttt{dummy\_classifier.py}.

\subsection{Implementation Details}
The experiments were conducted on a Lambda Labs Vector Machine, equipped with Threadripper Pro 3990X v4 @ 4.3GHz (64 cores), 128 GB DDR4 RAM and 2x NVIDIA GeForce RTX 3090 (24 GB VRAM each). The implementation was carried out in PyTorch, using pretrained weights from ImageNet. The dataset was split into training, validation, and test sets with a ratio of 65:15:20 for the virtual reality (VR) and 68:12:20 for the real-world (DR(eye)VE) datasets to accommodate differences in dataset sizes. We report the mean and standard deviation of the test accuracy resulting from 10 distinct splits of our datasets for each type of run. The training was performed using the SGD optimizer with an initial learning rate of 0.001. The learning rate was adjusted using a scheduler to ensure convergence. Models were trained for a maximum of 100 epochs, with early stopping based on the validation loss (20 epochs) to prevent overfitting. For all models, gaze data was not used during inference. Trained weights were instead imported into the vanilla ViT architecture for predictions on the test dataset.

\end{document}